\documentclass[11pt]{article}
\usepackage[utf8]{inputenc}
\usepackage{amsmath, amssymb}
\usepackage[a4paper, margin=1in]{geometry}
\usepackage{setspace}
\usepackage{newpxtext}
\usepackage{authblk}
\usepackage[hyperindex,breaklinks]{hyperref}
\usepackage{xcolor}

\title{AI and the Decentering of Disciplinary Creativity}

\author[1,2]{Eamon Duede\thanks{Email: eduede@purdue.edu}}
\affil[1]{Purdue University}
\affil[2]{Argonne National Laboratory}

\date{} 

\begin{document}

\maketitle

\begin{abstract}
    This paper examines the role of artificial intelligence in scientific problem-solving, with a focus on its implications for disciplinary creativity. Drawing on recent work in the philosophy of creativity, I distinguish between creative approaches and creative products, and introduce the concept of disciplinary creativity—the creative application of discipline-specific expertise to a valued problem within that field. Through two cases in mathematics, I show that while computation can extend disciplinary creativity, certain approaches involving AI can serve to displace it. This displacement has the potential to alter (and, perhaps, diminish) the value of scientific pursuit.
\end{abstract}
\section{Preliminaries}

Philosophy has a long and storied tradition of worrying about new technologies displacing human capacities. In the \textit{Phaedrus}, Socrates worries that the invention of and increasing reliance upon writing results in two significant epistemic displacements. Writing is believed to, first, displace memory by storing contents that would otherwise be remembered in static characters external to the mind. In doing this, writing encourages a shift in our means of recall from something active and within ourselves to something passive and external. Because we can consult what has been written, we no longer need to remember ---we can simply be reminded. Moreover, writing is thought to, in turn, displace genuine wisdom. Since recollection no longer relies upon something internal to and constitutive of wisdom, nor on something with which we can actively debate and interrogate, in referring to what has been written to inform us, we come to only appear wise. 

But, of course, writing also serves to massively extend our capacities beyond what is possible through the exercise of memory alone. The wisdom of Socrates, for instance, reaches and continues to actively engage us from the distant past, not through an oral tradition in the service of sustaining collective memory, but through external storage in writing. In a sense, then, writing serves both as a prosthetic and as an amputation.

Worries analogous to those of Socrates' concern with writing have attended the invention of and reliance upon computation in science. For instance, after the introduction of mechanical (as opposed to human) calculators in mathematics and physics in the mid-$20$\textsuperscript{th} century, many were concerned that the result would be a loss of so-called `number sense' in those who use them. This concern was likely well-founded. After all, Poincaré, von Neumann, Gauss, and Feynman have all been credited with remarkable contributions to mathematics and physics owing in large part to their tremendously fine numerical intuition, itself iteratively refined through a lifetime of obsessive internal calculation. 

More recently, philosophers and scientists have begun to wrestle with a set of epistemological concerns that arise from the use of forms of computation in science that are far more powerful than mere calculators. For instance, it has been argued that increasingly routine reliance on artificial intelligence leads scientists to adopt beliefs that are not fully justifiable due to the complexity and opacity of the models that support them. Moreover, it has been argued that the epistemic opacity of these systems limits scientific understanding of the phenomena under investigation, perhaps raising a dark veil between the practice of science and scientific knowledge. 

These concerns echo early $21$\textsuperscript{st} century worries surrounding the use of computational simulation in science. Notably, Paul Humphreys, in considering the impact of such computational approaches on scientific knowledge, argued that contemplation of these impacts is philosophically interesting precisely because the methods of computational science induce a displacement perhaps more general than that which exercised Socrates. For Humphreys, the hybrid scenario in which humans and machines work in concert to generate new scientific knowledge is thought to give rise to a novel displacement in which computationally assisted science pushes ``humans away from the centre of the epistemological enterprise.'' \cite[pg. 616]{humphreys2009philosophical}

A common thread that runs through this history of concerns is the sense that what is gained, whether by writing, using calculators, or some other technological advance, is accompanied by a loss. In the case of calculation, what is gained is the capability to very quickly solve or estimate solutions to otherwise intractable mathematical or statistical problems. But, what is lost is a feeling for numbers and functions that was, itself, of great utility for making mathematical progress more generally. Moreover, like in the case of writing, the thought is that what is lost or traded away is an ability or capacity of \textit{ours}. 

This paper joins the tradition and argues that, in addition to pushing humans away from the center of the \textit{epistemological} enterprise, recent advances in the use of artificial intelligence in science signal a concomitant displacement of \textit{disciplinary creativity} and expertise, and that this decentering has the potential to significantly scale up our collective problem-solving capacity by extending it beyond what is possible through the exercise of disciplinary creativity alone. But, this decentering also has the potential to radically alter (and, perhaps, diminish) the value of scientific pursuit.

Here, then, is the plan. In Section~\ref{creativity}, drawing from the literature on creativity, I bring out a distinction between creative approaches and creative products. Focusing on the latter, in Section~\ref{d_creativity} I develop the concept of disciplinary creativity and claim that, in the case of mathematics, we are likely to observe an increase in the stock of mathematically creative proofs (products), but the disciplinary creativity employed by the human in the generation of the proof (approaches) needn’t be specifically mathematical creativity.
In Section~\ref{comp_creativity}, I present two cases from mathematics involving computation in which researchers solve valued problems in ways that are arguably quite creative. However, only the first depends on mathematical disciplinary creativity.

\section{Creativity}
\label{creativity}

Philosophers have long been interested in creativity. Great works of art, literature, inventions of great utility, as well as scientific and philosophical theories that transform how we understand our world and place in it ---these all seem to display and depend upon acts of creativity. Philosophical interest has spawned a diverse literature containing a wide range of accounts of what constitutes creativity as a capacity. In trying to make sense of creativity as a capacity, Berys Gaut notes \cite{gaut2010philosophy} that in the \textit{Ion}, Plato likens creative states to a form of madness in which the mind and senses abandon us \cite{platoion}. Whereas, in the \textit{Critique of Judgment}, Kant links creativity to an exercise of the imagination unchecked by the understanding \cite{kant2000critique}. In contrast to accounts advancing something like a transcendent conception, in the \textit{Poetics}, Aristotle saw creativity as a skillful application of rules and heuristics \cite{halliwell1995aristotle}.

In addition to attempting to define creativity, contemporary work has sought to understand what distinguishes creative \textit{approaches} from creative \textit{results}. There is general agreement that we can judge some results (say a work of art or a scientific theory) as more creative than others. A prominent view in this respect is due to Boden who holds that results, whether an idea or artifact, are judged to be creative if they are novel, surprising, and valuable \cite{boden1998creativity, kind2022imagination}. More recently, debate has concerned the relationship between approaches and results, and, specifically, whether an approach can be deemed creative regardless of the status of its result \cite{hills2019against,langkau2024creative}. This debate turns on the relevance and importance of `value' in judgments concerning creativity with some arguing that, without a valued result, an approach cannot be deemed creative. For the purposes of this paper, we can set debates concerning value aside. While I am sympathetic to the idea that judgments concerning creativity need not involve ascriptions of value, it seems reasonable that there are particular contexts in which valuable solutions require creativity. That is, sometimes, while value may or may not required for creativity, creativity is required for value.

One area where this idea seems particularly salient is in science and, more specifically, scientific problem-solving. Reflecting on scientific problem-solving, Herbert Simon and Allen Newell argued that creative problem-solving is a distinctive kind of problem-solving generally, marked, in a manner similar to Boden's conception, by novelty, unconventional approaches, sustained effort, and challenges in framing the problem itself \cite{newell1962processes}. Boden further differentiates approaches into combinational (combining existing ideas), exploratory (extending existing conceptual frameworks), and transformational (radically altering conceptual spaces) creativity, each corresponding to different broad categories of creative approach \cite{boden1998creativity,boden2005creativity}. While there is no consensus on what features or characteristics an approach must posses to be deemed creative \cite{gaut2010philosophy}, there is general agreement that an approach will \textit{not} count as creative if it centrally depends on guessing, pure luck (knocking over paint buckets and producing beautiful images), a purely random process (chimps scribbling away), brute force (mechanically checking every possible solution) \cite{moruzzi2021measuring,deutsch1991creation,dutton2009art,gaut2010philosophy,halina2021insightful,hills2019against}\footnote{Though, as we will see in considering the proof plan for the four color theorem, \textit{some} brute-force approaches can still be hard-won and even ingenious}. As we will see in Section~\ref{d_creativity}, for any given problem, there may be many different approaches to its solution. Some problems can be solved through the application of either creative or uncreative approaches.

\section{Disciplinary Creativity}
\label{d_creativity}

Coming up with creative approaches to solving problems is indispensable for the sciences \cite{simonton2004creativity, dunbar1997scientists}. Yet, the sciences are made up of many distinct disciplines $\mathcal{D}$ the problems and approaches of which are highly varied. So, to fix ideas, let's consider one specific discipline ---mathematics. Suppose that mathematicians greatly value a solution to a particular problem $\mathcal{P}$ (say, the proof of some long-standing conjecture). There may be any number of reasons why mathematicians value a solution to $\mathcal{P}$. However, it is evident that they do so from the many observable attempts they have collectively made at solving it over the years \cite{venkatesh2024some}. Now, suppose someone solves the problem. From what we have seen already, determining whether the approach to the solution should be deemed \textit{creative} would depend on factors that are, perhaps, still to be worked out. But, most all accounts would agree that the solution would not count as a creative one if it were arrived at by, for instance, successive guessing, a purely random process, a brute force approach involving the mechanical checking of every possible solution, or by sheer luck.

As a result, in order for a solution to $\mathcal{P}$ to count as a \textit{creative} solution, it would need to meet the following condition:

\begin{quote}
    \textbf{Creative Approach}: $\mathcal{P}$ is solved by means of a specific approach that would be deemed creative on the correct account of creativity.
\end{quote}

For our purposes, the apparent circularity of \textbf{Creative Approach} is not problematic because our immediate goal is only to have a conceptual resource with which to distinguish between creative approaches and merely successful problem-solving approaches rather than to fully define creativity itself. In this way, we can black-box a complete specification of creativity (leaving the precise details deliberately unspecified) while still harvesting sufficient conceptual resources to, in what follows, construct a philosophically helpful definition of disciplinary creativity.

Now, for any given problem in science, there may be many solutions. For example, suppose we are confronted with the problem $\mathcal{P}$ of finding the sum of the first $100$ positive integers. Let us consider three approaches to solving this problem.

\textit{Approach One} to solving the problem is due to Gauss, who observed the following symmetry. If one pairs the first and last integers (1 and 100), the second and second-to-last (2 and 99), and so on, each pair sums to the same value:

\[(1 + 100) + (2 + 99) + (3 + 98) + \cdots + (50 + 51).\]

Since each pair sums to 101, and there are 50 such pairs, the solution to $\mathcal{P}$ is:

\[50 \times 101 = 5050.\]

\textit{Approach Two} to solving $\mathcal{P}$ exploits an observed geometric symmetry. Here, one can imagine dots arranged in a triangular pattern (as below). Each dot corresponds to an integer. One starts by placing a single dot in the first row, then two dots in the second row, three in the third, continuing in this way up to 100 dots in the 100th row.

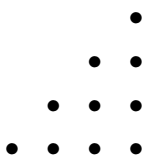
\[
\begin{array}{cccccccc}
        &       &       & \bullet \\
        &       & \bullet & \bullet \\
        & \bullet & \bullet & \bullet \\
\bullet & \bullet & \bullet & \bullet \\
\end{array}
\]

If one were to duplicate this triangle and rotate the copy, the two together would form a rectangle with dimensions 100 by 101. Of course, the area of the rectangle is the product of its dimensions and represents the total number of dots in both triangles. So, dividing the area by two yields the solution to $\mathcal{P}$:

\[
\frac{100 \times 101}{2} = 5050.
\]

There are many more solutions to $\mathcal{P}$, but let's consider a brute force, computational approach. \textit{Approach Three} to solving the problem involves using a computer. In \texttt{Python}, one can simply call the following:

\[\texttt{sum(range(1, 101))}\]

In this case, the computer begins by adding $0$ and $1$, then adds $2$ to that sum, then $3$ to the result, and so on, continuing in this way until it has added all of the integers from $1$ to $100$.\footnote{Note that in \texttt{Python}, the \texttt{range} function is half-open, meaning \texttt{range(1,\ 101)} includes $1$ but excludes $101$.} In doing so, Approach Three definitively solves $\mathcal{P}$ by brute force ---but without recognizing or exploiting any deeper structure in the problem.

Of our three approaches to solving $\mathcal{P}$, we can say with some certainty that Approach Three, in which we used a computer to sum the first $100$ integers, fails to satisfy the condition specified by Creative Approach because the approach is brute-force and purely mechanical in nature. Now, for the sake of argument, assume that Gauss' approach (Approach One) satisfies the conditions of Creative Approach. That is, on the correct account of creativity, Gauss' approach would count as creative. Since (on assumption) at least one of our approaches satisfies Creative Approach, and at least one does not, and since all three of the considered approaches led to a solution to the problem $\mathcal{P}$, it is clear that an approach to a problem need not satisfy the conditions on Creative Approach to, nevertheless, count as a solution to a problem. Importantly, however, to count as \textit{creative}, such conditions must be met. 

We are now in a position to put forward an account of \textit{disciplinary creativity} and identify conditions under which it may be identified. First, notice that Creative Approach is general enough to be applied across broadly heterogeneous problems and problem areas. For instance, the set of plausible approaches to the problem of designing a vaccine for a novel virus and the set of plausible approaches to the problem of finding the sum of the first $100$ positive integers are likely to be completely different. Nevertheless, there are creative and uncreative ways to accomplish both. Moreover, these distinct sets of plausible approaches do not differ merely in content but, rather, are shaped and constrained by the disciplinary frameworks within which the problems they address emerge. Vaccine design operates on expertise in and within the constraints of immunology, virology, biochemical engineering, and clinical trial protocols. Mathematical problem-solving employs expertise in and respects constraints of axioms, theorems, and obeys formal reasoning rules and structures unique to mathematics and logic. Finally, it is widely acknowledged that disciplinary boundedness indicates that creativity in each domain requires not just general creative capacity, but domain-specific expertise and skill \cite{weisberg2009out} that can be creatively (or uncreatively) applied to problems that are recognizable to and valued by its practitioners \cite{baer1998case, boden2005creativity, boden1998creativity}.

This suggests that an approach to a problem will demonstrate \textit{disciplinary} creativity only if it satisfies at least the following three conditions. The first is simply that the approach must meet the conditions of Creative Approach (e.g., it has to be \textit{creative}). The second is a valuation condition on the problem such that $\mathcal{P}$ is recognizable as a problem for a specific discipline $\mathcal{P_D}$ and is recognized as valuable, significant, or otherwise important to experts of that discipline $\mathcal{D}$ from which it emerges. This anchors the problem within a disciplinary context and indicates that its resolution bears significance for that discipline. The third is that the approach involves the appropriate application $\mathcal{A}$ of knowledge, skills, or methodologies distinctive to discipline $\mathcal{D}$. This condition indicates that the approach is centered on, draws upon, and is responsive to the norms, resources, and background knowledge of the relevant discipline.

From this, we can define disciplinary creativity in the following way:

\begin{quote}
    \textbf{Disciplinary Creativity}: the creative application $\mathcal{A}$ of discipline-$\mathcal{D}$-specific expertise to solve some problem $\mathcal{P_D}$ valued by experts in $\mathcal{D}$.    
\end{quote}

Consider Andrew Wiles' proof of Fermat's Last Theorem \cite{wiles1995modular,taylor1995ring}. This number-theoretic problem $\mathcal{P_D}$ was long recognized as significant and highly valued by mathematicians $\mathcal{D}$. It asks whether there exist positive integers \(a\), \(b\), and \(c\) such that \(a^n + b^n = c^n\) for any integer \(n > 2\). Wiles' approach involved the application of highly specialized discipline-specific tools and methods. Rather than attacking the problem directly, he proved a special case of the Modularity Theorem, establishing a deep and previously conjectural connection between elliptic curves and modular forms. Suppose we assume (and this is likely as safe an assumption as one can make) that Wiles' approach satisfies Creative Approach. Then, the resulting proof meets the conditions for Disciplinary Creativity: the creative application of discipline-\(\mathcal{D}\)-specific expertise to a problem \(\mathcal{P_D}\) recognized as valuable by experts in \(\mathcal{D}\).

By contrast, consider a hypothetical case in which a group of physicists attempts to design a vaccine for a novel virus, $\mathcal{P}$, using only tools and concepts from statistical mechanics, without drawing on established immunological knowledge, virological data, or clinical trial design. Suppose further that their approach satisfies the criteria for Creative Approach and ultimately produces a viable vaccine. Even so, it would fail to meet the conditions for Disciplinary Creativity. Although $\mathcal{P}_{n}$ is a problem recognized as valuable within a specific discipline $\mathcal{D}_{n}$ (namely, immunology or virology) the approach does not involve the application of expertise distinctive to that discipline. Instead, it represents the creative application of discipline-$\mathcal{D}_{m}$-specific expertise (physics) to a problem situated outside $\mathcal{D}_{m}$'s domain and addresses a problem not recognized as valuable by experts in $\mathcal{D}_{m}$. In this case, the result is still creative, but it does not qualify as disciplinary creativity. This contrast helps clarify what is at stake in identifying an approach as an instance of disciplinary creativity ---not merely that it is creative, but that it arises from, and contributes to, the epistemic life of a specific discipline.

While the hypothetical case of vaccine design is admittedly fanciful, it nonetheless illustrates how disciplinary creativity can be displaced in some problem space. The cases I introduce in Section~\ref{comp_creativity}, however, bring out how computation can both preserve and extend disciplinary creativity in mathematics and displace it.

\section{Computation and Disciplinary Creativity}
\label{comp_creativity}

\subsection{4CT: Mechanized Mathematical Creativity}


In 1976, mathematicians Kenneth Appel and Wolfgang Haken proved what is commonly referred to as the Four Color Theorem (4CT) \cite{appel1989every}. The theorem says that every planar graph (no loops) can be colored using at most four colors such that no two adjacent vertices share the same color. Versions of the conjecture have been under investigation since at least the mid-19$^{th}$ Century.

The proof begins by noting that any planar graph can be transformed into a triangulation (a graph in which every face is a triangle) by adding edges without affecting colorability. This reduces the problem to proving the four-colorability of maximal planar graphs. Since this is an infinite class, the proof cannot proceed by brute force. Appel and Haken instead demonstrate that every planar triangulation must contain at least one configuration from a finite, explicitly constructed set known as the unavoidable set. To identify this set, they employ the method of discharging, which draws an analogy to electrical charge. Using Euler’s formula, they assign an initial charge to each vertex based on its degree, such that the total charge across the graph is fixed (in fact, negative). Discharging rules then redistribute this charge locally across the graph, preserving the total but concentrating positive charge in certain small configurations. These configurations are exhaustively enumerated and shown to occur in every triangulation—that is, they are unavoidable. The final step is to verify, using computer-assisted methods, that each of these configurations is reducible—that is, cannot occur in a minimal counterexample to the theorem. Since every triangulation must contain one of these reducible configurations, no minimal counterexample can exist, and the Four Color Theorem follows.


Here, the problem of proving 4CT is clearly a problem $\mathcal{P}_{m}$ belonging to the discipline of mathematics $\mathcal{D}_{m}$. While the approach does involve a brute-force component, it remains disciplinarily mathematical. In the final step, Appel and Haken employed a supercomputer to verify the four-colorability of each configuration in the unavoidable set. This step is often characterized as brute-force, but this characterization is misleading. The machine is not exhaustively checking every possible coloring of every planar graph. Rather, the program is verifying the four-colorability of a finite class of configurations explicitly identified by prior steps in the proof. This verification could, in principle, be carried out by hand (albeit with great difficulty and requiring untold amounts of time). It has been noted that, in this way the machine is performing nothing other than a mechanized mobilization and application of ordinary human mathematical capacities  \cite{burge1998computer,duede2024apriori}. Moreover, because the proof proceeds by reducing the infinite class of graphs to a finite, representative set and establishing a general result by covering these cases, it follows the structure of a proof by induction. Far from undermining its mathematical standing, the use of computation in this context reflects the (potentially) creative application of discipline-$\mathcal{D}_{m}$-specific methods to a problem $\mathcal{P}_{m}$ recognized as central and valuable to mathematics.

In contrast to the Appel-Haken case, I turn now to a case involving the use of artificial intelligence in mathematics, the analysis of which illustrates a displacement of disciplinary creativity.

\subsection{Displaced Mathematical Creativity}
\label{displaced}

In the 4CT case, a computer encodes and carries out the mathematically creative approach conceived by mathematicians Appel and Haken to evaluating the four-colorability of planar graphs. In this section, I consider a case in which computational systems involving a large language model are arranged in a way that results in a breakthrough in a problem area of extremal combinatorics known as the Cap Set Problem. A \textit{cap set} is a subset of $(\mathbb{Z} / 3\mathbb{Z})^n$ containing no three distinct elements that sum to $0$ (\texttt{mod} $3$). For each dimension $n$, determining the maximum possible size of such a set remains an open challenge. While this number is necessarily bounded above by $3^n$ \cite{grochow19}, exact values have only been established for $n \leq 6$. Importantly, for our purposes, the solution space grows exponentially with increasing $n$. As a result, it is, in practice, not possible to make progress on the problem by means of brute-force computational approaches. This is the kind of problem for which it seems creativity is required.

In \cite{romera2023mathematical}, the problem that researchers seek to find a solution to is the construction of the largest cap set for $n=8$. This is a problem $\mathcal{P_D}$ in and valued by experts in extremal combinatorics $\mathcal{D}$. The researchers’ approach to the problem is to devise a multi-step, algorithmic \textit{framework} connecting multiple computational structures that collectively construct cap sets for arbitrary values of $n$. Importantly (as I discuss below), this same framework can be leveraged to solve arbitrary problems, for which a solution can take the form of a computer program. The approach yielded a cap set of 512, the largest verified set for $n=8$. Previously, the largest known cap set for $n=8$ was 496, making this result a meaningful contribution to a valued problem in the discipline.

The approach is a framework built up from the following elements. A fast-inference large language model built on \texttt{PaLM2} called \texttt{Codey} \cite{anil2023palm}. This LLM is fine-tuned on a publicly available corpus of computer code and has as its objective the generation of accurate code across various programming languages. Importantly, the model is not specifically fine-tuned on domain-$\mathcal{D}$-specific text, nor is it trained to solve problems in mathematics more generally. The framework also includes an evaluator called \texttt{evaluate()}, a program that checks proposed solutions to the problem and scores them. \texttt{evaluate()} is the framework's only component containing any domain-specific information. Specifically, \texttt{evaluate()} checks whether a proposed set of vectors satisfies the cap set constraint that no three vectors sum to zero in $(\mathbb{Z} / 3\mathbb{Z})^n$. Finally, there is a database in which candidate solutions to the problem are stored.

The framework works as follows. The researchers pass a skeleton Python program containing a function called \texttt{solve(n)} to the LLM, where \texttt{n} corresponds to the dimension for which the largest cap set is sought. The LLM then generates code for a function called \texttt{priority()}, which assigns a numerical score to each candidate vector in $(\mathbb{Z} / 3\mathbb{Z})^n$ that reflects the desirability of including that vector in the cap set. Once the new \texttt{priority()} function is composed into the complete program, it is executed via a wrapper function (typically called \texttt{main()}) which calls \texttt{solve(n)} and passes the resulting set to the evaluator. If the vectors included in the cap set satisfy the cap set constraint, the evaluator assigns a numerical value equal to the size of the proposed cap set for $n$. The program is then stored in the database. This cycle (from the input of a skeleton to the generation of a candidate function to the evaluation of the resulting program and storage of programs that do not violate the cap set constraint) constitutes a single pass of the algorithm. On subsequent passes, a number of high-scoring programs are sampled from subpopulations in the database and prepended to the input to the LLM, which is prompted to improve the \texttt{priority()} function in light of these prior examples. This whole process repeats on the order of $10^6$ times until no further progress is made. 

This recursive approach enables the framework to converge on increasingly effective solutions without human intervention into the domain-specific logic of the problem. As I argue below, this architecture realizes a form of problem-solving that is creative but no longer disciplinary in the traditional sense.


Here, the problem of constructing the largest cap set for $n=8$ is a problem $\mathcal{P}_{c}$ belonging to the discipline of extremal combinatorics $\mathcal{D}_{c}$. The \textit{approach}, however, does not centrally involve the application of discipline-$\mathcal{D}_{c}$-specific knowledge or methods by a domain expert. Instead, the approach to solving the problem involves the arrangement of computational components (most notably, a large language model not specifically trained on mathematical texts or for the solving of mathematical problems) to recursively generate, modify, and test \texttt{Python} programs against a mathematical constraint. The only component of the system that encodes any discipline-$\mathcal{D}_{c}$-specific content is the \texttt{evaluate()} evaluator, which simply checks whether a given output satisfies the cap set condition. The LLM is not guided by the mathematical insights of its creators, nor does it mechanically implement techniques distinctive to combinatorics in a way analogous to Appel and Haken's machine. Rather, the LLM generates code that scores well according to the evaluator, and the framework iterates without any reference to, or embedding in, the discipline’s background knowledge, methods, or normative practices. 

While the approach's iterative, feedback-guided refinement of its own outputs may well count as creative in a general sense or in, say, an engineering sense, the approach fails to meet the criteria for disciplinary creativity. Even if we assume that the approach is creative, we cannot say that it came about through the creative application of discipline-$\mathcal{D}_{c}$-specific expertise (e.g., the exercise of mathematical creativity). The approach to finding a solution using the framework discussed here is also not an instance of interdisciplinary creativity, where insights from multiple disciplines are brought into productive conversation, but rather a case in which disciplinary centering is bypassed altogether (as in the imagined case of vaccine design considered above). Nevertheless, the approach is likely creative on the best account of creativity and does result in a solution to $\mathcal{P}_{c}$ that is recognizable as a solution to the relevant community in extermal combinatorics. In this way, the success represents a displacement of disciplinary creativity rather than an instance of it.

\section{Discussion}
\label{discussion}
It is important to note that the very same framework discussed in Section~\ref{displaced} is a general framework that has been applied to completely different problems in mathematics. That is, the approach is not an approach to solving any particular problem at all. If we take seriously the idea that resources drawn on in the exercise of mathematical creativity vary from problem to problem, then an approach that is indifferent to the particularities of any given problem represents a departure from the computational exercise of disciplinary creativity observed in case of the 4CT. The result is an approach to problem-solving that is both powerful and detached. Like the many technological advances that have troubled philosophers in the past, it powerful in the sense that it is likely to vastly extend our capacities. Yet, unlike those many technological advances, it is likely to not only further push humans away from the center of the epistemological enterprise, but to simultaneously displace disciplinary creativity and expertise from the center of our scientific problem-solving strategies.

\bibliographystyle{alpha}
\bibliography{bibliography}

\newcommand{\etalchar}[1]{$^{#1}$}
\begin{thebibliography}{RPBN{\etalchar{+}}23}

\bibitem[ADF{\etalchar{+}}23]{anil2023palm}
Rohan Anil, Andrew~M Dai, Orhan Firat, Melvin Johnson, Dmitry Lepikhin, Alexandre Passos, Siamak Shakeri, Emanuel Taropa, Paige Bailey, Zhifeng Chen, et~al.
\newblock Palm 2 technical report.
\newblock {\em arXiv preprint arXiv:2305.10403}, 2023.

\bibitem[AH76]{appel1989every}
Kenneth~I Appel and Wolfgang Haken.
\newblock {\em Every planar map is four colorable}, volume~82.
\newblock American Mathematical Soc., 1976.

\bibitem[Bae98]{baer1998case}
John Baer.
\newblock The case for domain specificity of creativity.
\newblock {\em Creativity research journal}, 11(2):173--177, 1998.

\bibitem[Bod98]{boden1998creativity}
Margaret~A Boden.
\newblock Creativity and artificial intelligence.
\newblock {\em Artificial intelligence}, 103(1-2):347--356, 1998.

\bibitem[Bod05]{boden2005creativity}
Margaret~A Boden.
\newblock What is creativity?
\newblock In {\em Creativity in human evolution and prehistory}, pages 27--55. Routledge, 2005.

\bibitem[Bur98]{burge1998computer}
Tyler Burge.
\newblock Computer proof, apriori knowledge, and other minds: The sixth philosophical perspectives lecture.
\newblock {\em Philosophical perspectives}, 12:1--37, 1998.

\bibitem[DD]{duede2024apriori}
Eamon Duede and Kevin Davey.
\newblock Apriori knowledge in an era of computational opacity: The role of ai in mathematical discovery.
\newblock {\em Philosophy of Science (Forthcoming)}.

\bibitem[Deu91]{deutsch1991creation}
Harry Deutsch.
\newblock The creation problem.
\newblock {\em Topoi}, 10(2):209--225, 1991.

\bibitem[Dun97]{dunbar1997scientists}
Kevin Dunbar.
\newblock How scientists think: On-line creativity and conceptual change in science.
\newblock 1997.

\bibitem[Dut09]{dutton2009art}
Denis Dutton.
\newblock {\em The art instinct: Beauty, pleasure, \& human evolution}.
\newblock Oxford University Press, USA, 2009.

\bibitem[Gau10]{gaut2010philosophy}
Berys Gaut.
\newblock The philosophy of creativity.
\newblock {\em Philosophy Compass}, 5(12):1034--1046, 2010.

\bibitem[Gro19]{grochow19}
Joshua~A. Grochow.
\newblock New applications of the polynomial method: the cap set conjecture and beyond.
\newblock {\em Bull. Amer. Math. Soc.}, 56(1):29--64, 2019.

\bibitem[Hal95]{halliwell1995aristotle}
Francis~Stephen Halliwell.
\newblock {\em Aristotle Poetics (with Longinus on the Sublime and Demetrius On Style)}.
\newblock Harvard University Press, 1995.

\bibitem[Hal21]{halina2021insightful}
Marta Halina.
\newblock Insightful artificial intelligence.
\newblock {\em Mind \& Language}, 36(2):315--329, 2021.

\bibitem[HB19]{hills2019against}
Alison Hills and Alexander Bird.
\newblock Against creativity.
\newblock {\em Philosophy and Phenomenological Research}, 99(3):694--713, 2019.

\bibitem[Hum09]{humphreys2009philosophical}
Paul Humphreys.
\newblock The philosophical novelty of computer simulation methods.
\newblock {\em Synthese}, 169:615--626, 2009.

\bibitem[Kan00]{kant2000critique}
Immanuel Kant.
\newblock {\em Critique of the Power of Judgment}.
\newblock The Cambridge Edition of the Works of Immanuel Kant. Cambridge University Press, 2000.

\bibitem[Kin22]{kind2022imagination}
Amy Kind.
\newblock {\em Imagination and creative thinking}.
\newblock Cambridge University Press, 2022.

\bibitem[Lan24]{langkau2024creative}
Julia Langkau.
\newblock What is creative imagining?
\newblock {\em Analysis}, page anae059, 2024.

\bibitem[Mor21]{moruzzi2021measuring}
Caterina Moruzzi.
\newblock Measuring creativity: an account of natural and artificial creativity.
\newblock {\em European Journal for Philosophy of Science}, 11(1):1, 2021.

\bibitem[NSS62]{newell1962processes}
Allen Newell, J~Clifford Shaw, and Herbert~A Simon.
\newblock The processes of creative thinking.
\newblock In {\em Contemporary Approaches to Creative Thinking.} Atherton Press, 1962.

\bibitem[Pla61]{platoion}
Plato.
\newblock Ion.
\newblock In Edith Hamilton and Huntington Cairns, editors, {\em The Collected Dialogues of Plato}. Princeton University Press, 1961.

\bibitem[RPBN{\etalchar{+}}23]{romera2023mathematical}
Bernardino Romera-Paredes, Mohammadamin Barekatain, Alexander Novikov, Matej Balog, M~Pawan Kumar, Emilien Dupont, Francisco~JR Ruiz, Jordan~S Ellenberg, Pengming Wang, Omar Fawzi, et~al.
\newblock Mathematical discoveries from program search with large language models.
\newblock {\em Nature}, pages 1--3, 2023.

\bibitem[Sim04]{simonton2004creativity}
Dean~Keith Simonton.
\newblock {\em Creativity in science: Chance, logic, genius, and zeitgeist}.
\newblock Cambridge University Press, 2004.

\bibitem[TW95]{taylor1995ring}
Richard Taylor and Andrew Wiles.
\newblock Ring-theoretic properties of certain hecke algebras.
\newblock {\em Annals of Mathematics}, 141(3):553--572, 1995.

\bibitem[Ven24]{venkatesh2024some}
Akshay Venkatesh.
\newblock Some thoughts on automation and mathematical research.
\newblock {\em Bull. Amer. Math. Soc.(NS)}, 61(2):203--210, 2024.

\bibitem[Wil95]{wiles1995modular}
Andrew Wiles.
\newblock Modular elliptic curves and fermat's last theorem.
\newblock {\em Annals of mathematics}, 141(3):443--551, 1995.

\bibitem[WM09]{weisberg2009out}
Robert~W Weisberg and A~Markman.
\newblock On “out-of-the-box” thinking in creativity.
\newblock {\em Tools for innovation}, pages 23--47, 2009.

\end{thebibliography}

\end{document}